\documentclass[sigconf]{acmart}

\usepackage{times}
\usepackage{epsfig}
\usepackage{graphicx}
\usepackage{amsmath}

\usepackage{multirow}%合并行
\usepackage{amssymb}%不规则表格
\usepackage{geometry}%不规则表格
\usepackage{booktabs}%不规则表格
\usepackage{diagbox}%不规则表格
\usepackage{array}%字符水平居中

\usepackage{algorithm}%算法
\usepackage{algorithmic}%算法

\usepackage{multirow}

\AtBeginDocument{%
  }

\setcopyright{acmcopyright}
\copyrightyear{2023}
\acmYear{2023}
\acmDOI{XXXXXXX.XXXXXXX}

\acmSubmissionID{56}
\renewcommand\footnotetextcopyrightpermission[1]{}%这个是首页的尾注，加上这段代码则不显示，录用后去掉这段代码
\begin{document}

\title{Adversarial Neon Beam: A Light-based Physical Attack to DNNs}

% \affiliation{%

\author{Chengyin Hu}
\affiliation{%
  \institution{University of Electronic Science and Technology of China}
  % \streetaddress{1 Th{\o}rv{\"a}ld Circle}
  \city{Chengdu}
  \country{China}}
\email{cyhuuesct@gmail.com}

\author{Weiwen Shi}
\affiliation{%
  \institution{University of Electronic Science and Technology of China}
  % \streetaddress{1 Th{\o}rv{\"a}ld Circle}
  \city{Chengdu}
  \country{China}}
\email{weiwen_shi@foxmail.com}

\author{Wen Li}
\authornote{Corresponding authors}
\affiliation{%
  \institution{University of Electronic Science and Technology of China}
  % \streetaddress{1 Th{\o}rv{\"a}ld Circle}
  \city{Chengdu}
  \country{China}}
\email{liwen@uestc.edu.cn}

% \affiliation{%

% \affiliation{%

% \affiliation{%

% \affiliation{%

% \affiliation{%

% \affiliation{%


\begin{abstract}
 In the physical world, deep neural networks (DNNs) are impacted by light and shadow, which can have a significant effect on their performance. While stickers have traditionally been used as perturbations in most physical attacks, their perturbations can often be easily detected. To address this, some studies have explored the use of light-based perturbations, such as lasers or projectors, to generate more subtle perturbations, which are artificial rather than natural. In this study, we introduce a novel light-based attack called the adversarial neon beam (\textbf{AdvNB}), which utilizes common neon beams to create a natural black-box physical attack. Our approach is evaluated on three key criteria: effectiveness, stealthiness, and robustness. Quantitative results obtained in simulated environments demonstrate the effectiveness of the proposed method, and in physical scenarios, we achieve an attack success rate of 81.82\%, surpassing the baseline. By using common neon beams as perturbations, we enhance the stealthiness of the proposed attack, enabling physical samples to appear more natural. Moreover, we validate the robustness of our approach by successfully attacking advanced DNNs with a success rate of over 75\% in all cases. We also discuss defense strategies against the AdvNB attack and put forward other light-based physical attacks. %Our findings showcase the potential of light-based attacks in enhancing the effectiveness, stealthiness and robustness of physical attacks to DNNs.
\end{abstract}


%    <concept>
%        <concept_id>10002978.10003029.10003032</concept_id>
%        <concept_desc>Security and privacy~Social aspects of security and privacy</concept_desc>
%        <concept_significance>300</concept_significance>
%        </concept>
%  </ccs2012>

% \ccsdesc[300]{Security and privacy~Social aspects of security and privacy}


\maketitle

\section{Introduction}
\label{sec1}

Deep learning has demonstrated remarkable achievements in various fields, even surpassing human-level performance \cite{ref55,ref56,ref57}. As a result, it has been extensively employed in applications such as autonomous vehicles \cite{ref50, ref51}, robotics \cite{ref52}, and UAVs \cite{ref53,ref54}. However, deep neural networks (DNNs) are vulnerable to well-crafted adversarial perturbations \cite{ref1,ref2,ref3}, leading to unpredictable aberrant behavior in vision-based systems. For instance, in autonomous driving, an attack on DNNs could result in car crashes. Adversarial attacks have mainly been investigated in the digital domain \cite{ref59,ref61,ref64,ref67}, where subtle perturbations are intentionally added to input images, making them difficult for human observers to detect. Recently, some researchers have focused on physical adversarial attacks \cite{ref71,ref90,ref92,ref75}, which involve capturing images by a camera and then feeding them to the target model. Physical perturbations are typically designed to be visible so that they can be captured by the camera, yet at the same time, they need to be inconspicuous enough to avoid detection by human observers. Therefore, physical attacks typically involve a trade-off between stealthiness and robustness.

\begin{figure}
\centering
\includegraphics[width=1\columnwidth]{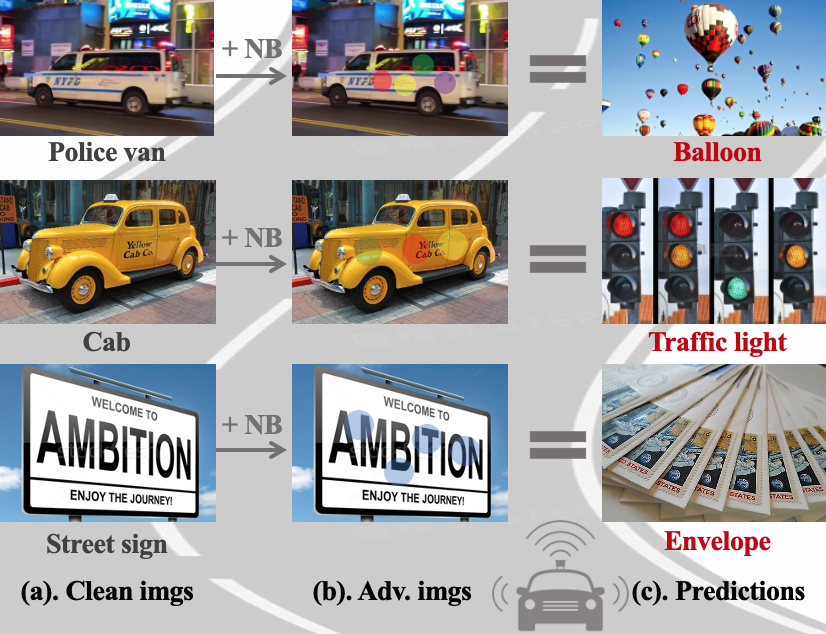} 
\caption{An example. 
When the camera of a self-driving system captures objects illuminated by neon beams, it may fail to recognize them accurately. }.
\vspace{-0.5cm}
\label{figure1}
\end{figure}

Numerous natural phenomena can serve as physical perturbations. In urban areas, in particular, neon beams are ubiquitous and often scatter onto traffic signs, leading humans to unconsciously disregard them. However, if an attacker intentionally generates adversarial neon beams to fool self-driving car systems while lowering human vigilance, it could severely disrupt traffic flow. As illustrated in Figure \ref{figure1}, an adversary can project ingeniously crafted adversarial neon beams onto a road sign, leading advanced DNNs to misclassify it.

\begin{figure}
\centering
\includegraphics[width=0.7\columnwidth]{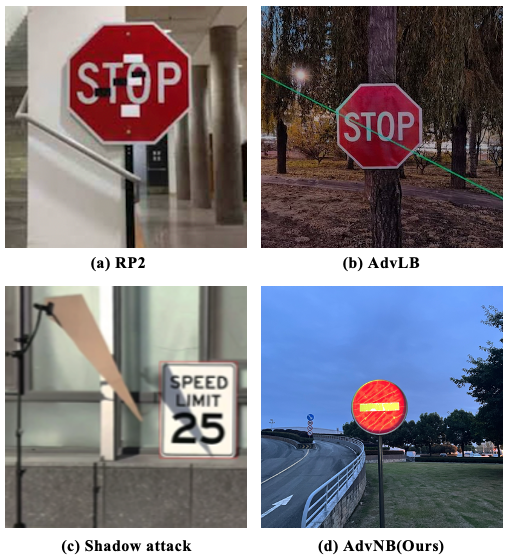} 
\caption{Visual comparison. }.
\vspace{-0.5cm}
\label{figure2}
\end{figure}

The investigation of the adversarial impact of light-based physical attacks on advanced DNNs is of great significance. Currently, most physical attacks employ stickers as perturbations \cite{ref24,ref26}, which successfully deceive advanced DNNs without altering the semantic information of the target objects. However, sticker-based attacks entail affixing stickers or posters onto the surface of the target object, which lacks flexibility and makes it difficult to achieve stealthiness. Additionally, some researchers have examined camera-based adversarial attacks \cite{ref38},in which some tiny translucent patches are placed on the camera of a mobile phone to perform physical attacks. However, adding too many patches can result in experimental errors. Recently, some scholars have used light beams as physical perturbations \cite{ref35,ref78} to conduct instantaneous attacks, which effectively deceive advanced DNNs and offer better stealthiness. However, this method is prone to failure during daylight hours.

In this work, we present a light-based physical attack called Adversarial Neon Beam (AdvNB). Unlike existing sticker-based attacks, our method employs the instantaneous attack nature of neon beams to execute physical attacks, rendering it more flexible and stealthy. Compared to other light-based methods, our approach is more natural and offers enhanced stealthiness. A visual comparison of our method with other works is depicted in Figure \ref{figure2}. The adversarial samples generated by AdvNB are significantly stealthier than those generated by other methods such as RP2 \cite{ref24}, AdvLB \cite{ref35}, and shadow attack \cite{ref37}.

% The physical perturbations generated by AdvNB look like normal neon beams in the busy city. Though the adversarial sample generated by AdvNB may appear same stealthiness with RP2 and AdvLB, AdvNB exhibits some advantages. e.g., The instantaneous attack nature of AdvNB show flexible operations than RP2 (See RP2 in Figure \ref{figure2}), the robustness of AdvNB is better than AdvLB (See section \ref{sec4}).

Our proposed method offers a straightforward implementation for physical attacks. We formalize the physical parameters of the neon beam, including the radius, intensity, color, and center position. Then, we design an optimization method to search for the most adversarial physical parameters. To achieve the transition from the digital domain to the physical domain, we employ EOT \cite{ref31}. Finally, based on these physical parameters, we project physical neon beams onto the target objects to generate physical samples. Notably, our approach enables low-cost attacks, requiring a budget of less than 50 USD, which significantly facilitates deployment. Our main contributions can be summarized as follows:

\begin{itemize}
\item We introduce a novel light-based physical attack method, AdvNB, that exploits the instantaneous nature of neon beams to manipulate their physical parameters and generate black-box physical attacks (see section \ref{sec1}). Our approach is effective and low-cost, making it significantly easier to deploy.% However, this ease and convenience also makes AdvNB a potential safety threat.
\item We introduce and analyze the existing methods (See section \ref{sec2}), analyze the advantages of our approach over traditional road sign attacks and the existing light-based attacks. Then, we design strict experimental method and conduct comprehensive experiments to verify the effectiveness, stealthiness and robustness of AdvNB (See section \ref{sec3}, \ref{sec4}). Our results indicate that AdvNB is capable of achieving high success rates while remaining inconspicuous to the naked eye, even in challenging real-world scenarios. Given these findings, we believe that AdvNB represents a valuable tool for further studying the threat posed by light-based attacks in realistic settings.
\item We perform a comprehensive analysis of AdvNB, which includes studying the impact of AdvNB on DNNs' prediction errors, and investigating the defense strategies against AdvNB (as detailed in section \ref{sec5}). These investigations serve to facilitate scholars in exploring and defending against light-based physical attacks. Moreover, we also examine some potential avenues for future research on light-based physical attacks (as discussed in section \ref{sec6}).
\end{itemize}

\section{Related work}
\label{sec2}

\subsection{Digital attacks}
The concept of adversarial attacks was initially introduced by Szegedy et al. \cite{ref1}, and subsequently, many digital attacks have been proposed \cite{ref16,ref17,ref20,ref21}.

Currently, many digital attacks aim to ensure that perturbations are imperceptible to human observers by confining them to a norm-ball, with ${L}_{2}$ and ${L}_{\infty}$ being the most frequently employed norms \cite{ref85,ref86}. These methods have proven to be effective in attacking advanced deep neural networks while maintaining the perturbations' imperceptibility to humans. Other works have focused on modifying other characteristics of clean samples for adversarial attacks, such as color \cite{ref7,ref8,ref9}, texture, and camouflage \cite{ref87,ref11,ref88,ref13}, which are often noticeable to the naked eye. Additionally, some research has generated adversarial samples by manipulating the physical parameters of clean images, while retaining the critical components of the images, thus facilitating digital attacks \cite{ref14,ref15}. Several works \cite{ref79, ref80} have proposed the raindrop attack, which employs simulated raindrops as perturbations, to test its attack effectiveness and subsequently develop defense mechanisms for obtaining robust deep neural networks. In contrast to digital attacks, physical attacks cannot directly manipulate input images.

\begin{figure*}
\centering
\includegraphics[width=1\linewidth]{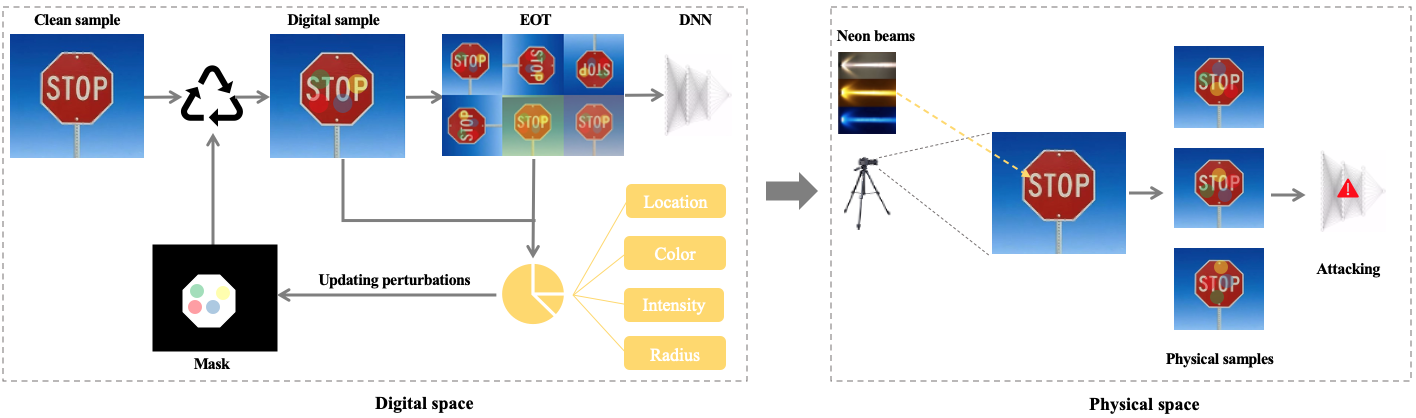}
\caption{Generating adversarial samples.}.
\label{figure3}
\end{figure*}

\subsection{Physical attacks}
Physical attack was first proposed by Alexey Kurakin et al. \cite{ref22}. After this work, many physical attacks were proposed successively \cite{ref89,ref28,ref31, ref91}.

\textbf{Traditional street sign attacks.} Ivan Evtimov et al. \cite{ref24} proposed RP2, which demonstrated a robust adversarial effect against advanced deep neural networks. However, RP2 is vulnerable to environmental interference at large distances and angles. Eykholt et al. \cite{ref26} improved RP2 and employed the attack on a target detector, resulting in the detector disregarding the target object. However, the perturbations cover an extensive area, they are too conspicuous. Chen et al. \cite{ref23} proposed ShapeShifter, which overcame the non-differentiability of the Faster R-CNN model, and successfully executed optimization-based attacks using gradient descent and back propagation. Huang et al. \cite{ref27} further enhanced ShapeShifter by adding Gaussian white noise to its optimization function, addressing ShapeShifter's high requirements for photographic equipment. However, both ShapeShifter and its improved version have a limitation: the perturbations cover almost the entire road sign, making them non-stealthy. Duan et al. \cite{ref25} proposed AdvCam, which leverages style transfer techniques to generate adversarial samples and disguise perturbations as a style that humans would consider reasonable. However, AdvCam requires manual selection of the attack area and target and suffers from printing losses. In general, traditional road sign attacks have significant drawbacks: adding physical perturbations is a manual and time-consuming process, while also susceptible to printing errors.

\textbf{Light-based attacks.} Light-based attacks have shown some advantages over traditional street sign attacks. For instance, Duan et al. \cite{ref35} proposed AdvLB, which allows for manipulations of the physical parameters of laser beams to perform physical attacks. However, AdvLB is prone to spatial errors. To address this limitation, Gnanasambandam et al. \cite{ref36} proposed OPAD, which projects digital perturbations onto the target objects to perform physical attacks on advanced DNNs. However, OPAD is prone to paralyzation during daytime conditions. Another physical attack approach is shadow-based physical attack, as studied by Zhong et al. \cite{ref37}, who use carefully crafted shadows as physical perturbations to generate physical samples and conduct effective attacks on advanced DNNs. However, the cardboard used in this approach is placed too close to the target objects, which may raise suspicions among human observers.

\textbf{Raindrop-based attacks.} Additionally, Guesmi et al. \cite{ref81} proposed Advrain, which performs physical attacks by simulating real raindrops. They achieved an average model accuracy reduction of 45\% for VGG19 and 40\% for ResNet34 using 20 raindrops. AdvRain implements stealthy physical attacks, but its robustness still needs to be explored.

\section{Approach}
\label{sec3}
% \subsection{Adversarial sample}
Given an input image $X$, a ground truth label $Y$, and a DNN classifier $f$, $f(X)$ represents the classifier's prediction label, the classifier $f$ associates with a confidence score ${f}_{Y}(X)$ to class $Y$. Generating adversarial sample ${X}_{adv}$ satisfies two properties : (1) $f({X}_{adv}) \neq f(X) = Y$; (2) $\parallel {X}_{adv} - X \parallel < \epsilon$. The first one requires that ${X}_{adv}$ successfully fool DNN classifier $f$, and the second one ensures the adversarial perturbations to be imperceptible to human observers.

Figure \ref{figure3} illustrates our proposed method, AdvNB. Firstly, we simulate neon beams and synthesize them with clean images to generate digital samples. Then, we use EOT \cite{ref31} to transition from the digital domain to the physical domain, followed by projecting the physical neon beams onto the target object to generate physical samples.

% \begin{figure*}
% \centering
% % \setlength{\belowcaptionskip}{-0.2cm}
% \includegraphics[width=1\linewidth]{figures/fig5.png}
% \caption{Adversarial samples generated by AdvNB.}.
% \label{figure5}
% \end{figure*}

% \begin{figure}
% \centering
% % \setlength{\belowcaptionskip}{-0.2cm}
% \includegraphics[width=1\columnwidth]{figures/fig6.png}
% \caption{Statistics on misclassification of adversarial samples. Through statistical analysis of all the misclassification results, we can see the semantic information categories of neon beams, among which the Shower curtain, Mosquito net and Envelope are the most contained.}.
% \label{figure6}
% \end{figure}

\subsection{Generating adversarial sample}
In our study, we define a neon beam by considering four key parameters: center position denoted by $L (m, n)$, radius represented by $R$, intensity denoted by $I$, and color represented by $C (r, g, b)$. Each parameter is defined as follows:

\textbf{Position $L (m, n)$:} We define the center position of the neon beam as $L (m, n)$. The neon beam is assumed to be circular and represented by a double tuple $(m, n)$ to indicate its center position.

\textbf{Radius $R$:} $R$ represents the radius of neon beam.

\textbf{Beam intensity $I$:} The parameter $I$ determines the strength of the neon beam projected onto the target object, with larger values indicating greater intensity and brighter illumination.

% we conditionally set the size of $I$ (values range from 0.1 to 0.7). In the physical environment, due to the limitation of the physical lighting equipment, we set the value range of $I$ to be from 0.2 to 0.6.

\textbf{Colors $C (r, g, b)$:} The parameter $C (r, g, b)$ determines the color of the neon beam. Here, $r$, $g$, and $b$ represent the intensity of the red, green, and blue channels of a digital image, respectively. In the physical environment, due to equipment limitations, we have chosen to use five common colors: Red $C(225,0,0)$, Green $C(0,255,0)$, Blue $C(0,0,255)$, Yellow $C(255,255,0)$, and Purple $C(255,0,255)$.

%$g$ represents the green channel, and $b$ represents the blue channel. In the physical environment, due to equipment limitations, we choose five common colors, including $Red (225,0,0)$; $Green (0,255,0)$; $Blue (0,0,255)$; $Yellow (255,255,0)$; $Purple (255,0,255)$.

The aforementioned parameters collectively define a neon beam ${\theta} (L, R, I, C)$, and the neon beams are grouped as ${\mathcal{G}}_{\theta}$, with each parameter having an adjustable range. $\mathcal{M}$ is defined as the mask to locate the target objects, and the neon beams present in the target objects are given by $\mathcal{M} \cap {\mathcal{G}}_{\theta}$. The proposed AdvNB aims to identify the neon beam group ${\mathcal{G}}_{\theta}$ that causes the simulated adversarial sample ${X}_{adv}$ to be misclassified by the target model $f$. To constrain the physical parameters $L$, $R$, $I$, and $C$ within appropriate ranges, we define the restriction vectors ${\vartheta}_{min}$ and ${\vartheta}_{max}$, which are adjustable. The adversarial example generation process can be denoted as:

% We define a function $Syn (X, {\mathcal{G}}_{\theta})$ to simply synthesize neon beam group ${G}_{\theta}$ with clean sample $X$ to generate adversarial sample ${X}_{adv}$:

\begin{equation}
    \label{Formula 1}
    {X}_{adv} = S(X, {\mathcal{G}}_{\theta}, \mathcal{M}) \quad \theta \in ({\vartheta}_{min},{\vartheta}_{max})
\end{equation}
% $$s.t. \quad \theta \in ({\vartheta}_{min},{\vartheta}_{max})$$
where ${X}_{adv}$ denotes adversarial samples and $S$ denotes simple linear fusion method to fuse clean sample $X$ and ${\mathcal{G}}_{\theta}$.

% $$s.t. \quad R \in [{\gamma}_{min},{\gamma}_{max}], I \in [{I}_{min},{I}_{max}]$$

% In the physical environment, in order to prevent neon beams from appearing in the background, we define a function $l$, to ensure physical neon beams project on the target objects. So, adversarial samples in the physical-setting can be expressed as follows:

% \begin{equation}
%     % \setlength{\abovedisplayskip}{3pt}
%     % \setlength{\belowdisplayskip}{3pt}
%     \label{Formula 2}
%     {X}_{adv} = Syn(X, l({G}_{\theta})) 
% \end{equation}

\textbf{Expectation Over Transformation.} In order to launch a successful physical-world attack, it is necessary to transform the digital samples into physical samples. Expectation Over Transformation (EOT) \cite{ref31} is a widely used method for this purpose, which ensures the generation of robust physical samples that can withstand various transformations, such as different distances and angles. We define a transformation function $\mathcal{T}$ to represent the domain transition, which is a random combination of digital image processing techniques, such as brightness adaptation, position offset, and color variation. By applying EOT, the physical sample can be expressed as:

\begin{equation}
    \label{Formula 3}
    {X}_{phy} = \mathcal{T}({X}_{adv}, {\mathcal{G}}_{\theta}) \quad \theta \in ({\vartheta}_{min},{\vartheta}_{max})
\end{equation}

% $$s.t. \quad \theta \in ({\vartheta}_{min},{\vartheta}_{max})$$

% We apply function \ref{Formula 3} to AdvNB for physical adaptation. The operations include: Adjusting physical parameters $L, R, I, C$, Adding random noise, etc. 

\begin{algorithm}
	\renewcommand{\algorithmicrequire}{\textbf{Input:}}
	\renewcommand{\algorithmicensure}{\textbf{Output:}}
	\caption{Pseudocode of AdvNB}
	\label{algorithm1}
	\begin{algorithmic}[1]
	
		\REQUIRE Input $X$, Ground truth label $Y$, Max step ${t}_{max}$, The number of neon beams $N$, Classifier $f$;
		\ENSURE A vector of parameters ${\mathcal{G}}_{\theta}$;

		\STATE \textbf{Initialization} ${\mathcal{G}}_{\theta}=\emptyset$, $\theta=\emptyset$, ${\theta}_{opt}=\emptyset$;  
		
		\STATE ${Score}^{\star}\leftarrow{f}_{Y}(X)$;
		
		\FOR{$i$ $\leftarrow$ 0 to $N$}
	        
	        \FOR{step $\leftarrow$ 0 to ${t}_{max}$}
	            \STATE Randomly pick $L, R, I, C$;
	            \STATE $\theta = {\theta}(L,R,I,C)$;
	            \STATE ${X}_{adv} = S({X}_{adv}, \theta, \mathcal{M})$;
	            \STATE ${f}_{Y}({X}_{adv}) \leftarrow f({X}_{adv})$;
	            
	            \IF{${Score}^{\star}>{f}_{Y}({X}_{adv})$}
	            \STATE ${Score}^{\star} \leftarrow {f}_{Y}({X}_{adv})$;
	            \STATE ${\theta}_{opt}=\theta$;
	            \ENDIF

	            % \IF{$Adjust(f({X}_{adv}); Y)$}

	            % \ENDIF

	        \ENDFOR
        \STATE ${\mathcal{G}}_{\theta} \leftarrow {\mathcal{G}}_{\theta}+{\theta}_{opt}$;
	    \STATE$ {X}_{adv} = S(X, {\mathcal{G}}_{\theta}, \mathcal{M})$;
        
        \ENDFOR
        \IF{$argmaxf({X}_{adv}) \neq argmaxf(X)$}
        \STATE \textbf{Return} ${\mathcal{G}}_{\theta}$;
	    % \STATE Exit();
        \ENDIF

	\end{algorithmic}  
\end{algorithm}

\subsection{Neon beam adversarial attack}
AdvNB focuses on searching ${\mathcal{G}}_{\theta}$, the physical parameters of adversarial neon beams, which generates an adversarial sample ${X}_{adv}$ that fools the target model $f$. In our test, we consider an attacker cannot attain the knowledge of the target model but only the confidence score ${f}_{Y}(X)$ with given input image $X$ on the ground truth label $Y$. In our proposed method, we use confidence score as the adversarial loss. Thus, the objective is formalized as minimizing the confidence score on the ground truth label $Y$, shown as follows:

\begin{equation}
    \label{Formula 4}
    \mathop{\arg\min}_{{\mathcal{G}}_{\theta}}{\mathbb{E}}_{t \sim \mathcal{T}}[{f}_{Y}(t({X}_{adv},{\mathcal{G}}_{\theta}))]
\end{equation}
$$s.t. \quad f({X}_{adv}) \neq Y$$

AdvNB can be divided into two main parts: the digital-setting and the physical-setting. In the digital-setting, we generate adversarial samples by randomly generating adversarial neon beams with parameters $L$, $R$, $I$ and $C$. In the physical-setting, we search for the most adversarial neon beams ${\mathcal{G}}_{\theta} = \{{\theta}_{1}, {\theta}_{2}, ..., {\theta}_{N}\}$ using the confidence score ${f}_{Y}(X)$ and generate physical samples with function \ref{Formula 3}.

Algorithm \ref{algorithm1} takes an input image $X$, ground truth label $Y$, a maximum number of iterations ${t}_{max}$, a maximum number of neon beams $N$, and a classifier $f$ as input. At each iteration, the algorithm searches for the most adversarial neon beam $\theta(L, R, I, C)$ that generates an adversarial sample with the lowest confidence score ${f}_{Y}({X}_{adv})$ on the ground truth label $Y$. It is important to note that a lower confidence score on the correct label of a digital sample indicates a more adversarial sample. The search terminates when the current ${X}_{adv}$ is predicted with a label $argmaxf({X}_{adv}) \neq argmaxf(X)$ or the maximum number of iterations ${t}_{max}$ is reached. Finally, the algorithm outputs the physical parameters of the neon beams ${\mathcal{G}}_{\theta}$, which are used to perform subsequent physical attacks.

\begin{figure}
\centering
\includegraphics[width=0.8\linewidth]{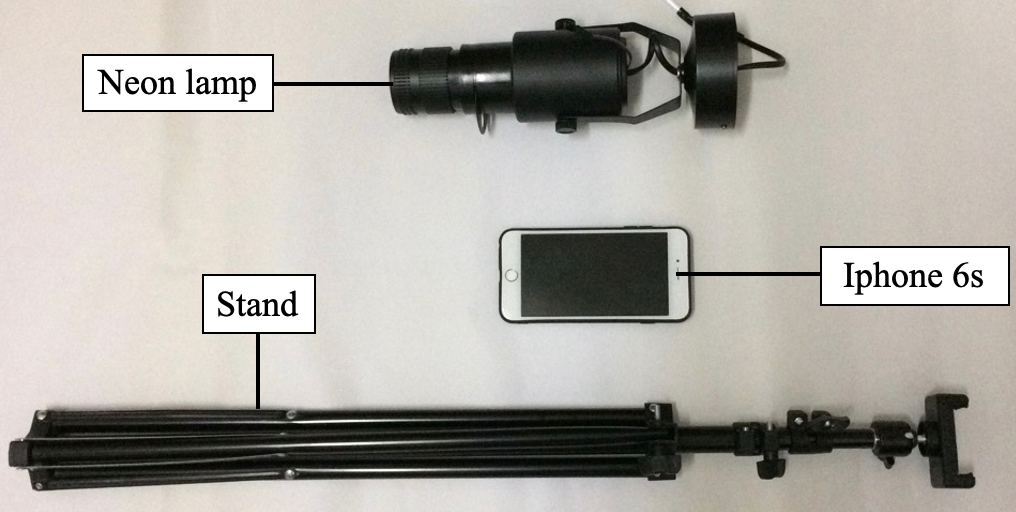}
\caption{Experimental devices. }.
\vspace{-0.5cm}
\label{figure4}
\end{figure}

\begin{figure*}
\centering
\includegraphics[width=1\linewidth]{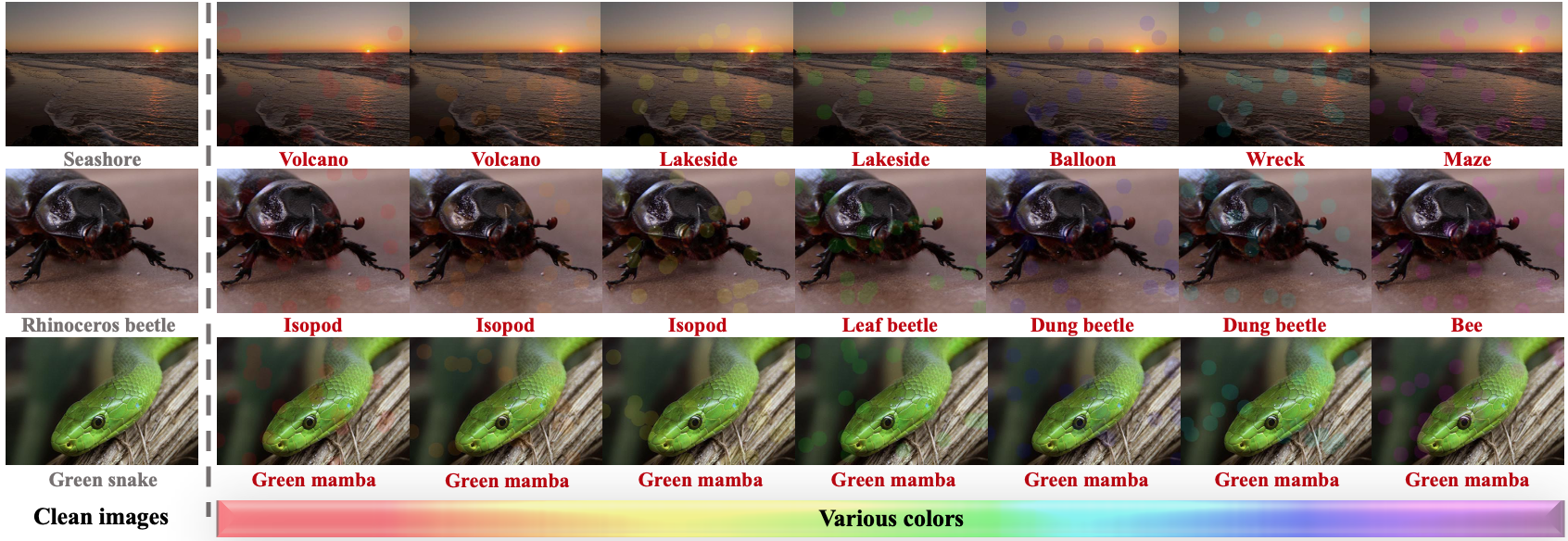}
\caption{Adversarial samples generated by AdvNB. To validate the adversarial effectiveness of neon beams in the digital environment, we conduct experiments using neon beams of various colors, including red, orange, yellow, green, blue, indigo, and purple, to attack target model $f$. The results demonstrate that the target model failed to correctly classify digital samples with neon beams of different colors added.}.
\label{figure5}
\end{figure*}

\begin{figure}
\centering
\includegraphics[width=1\columnwidth]{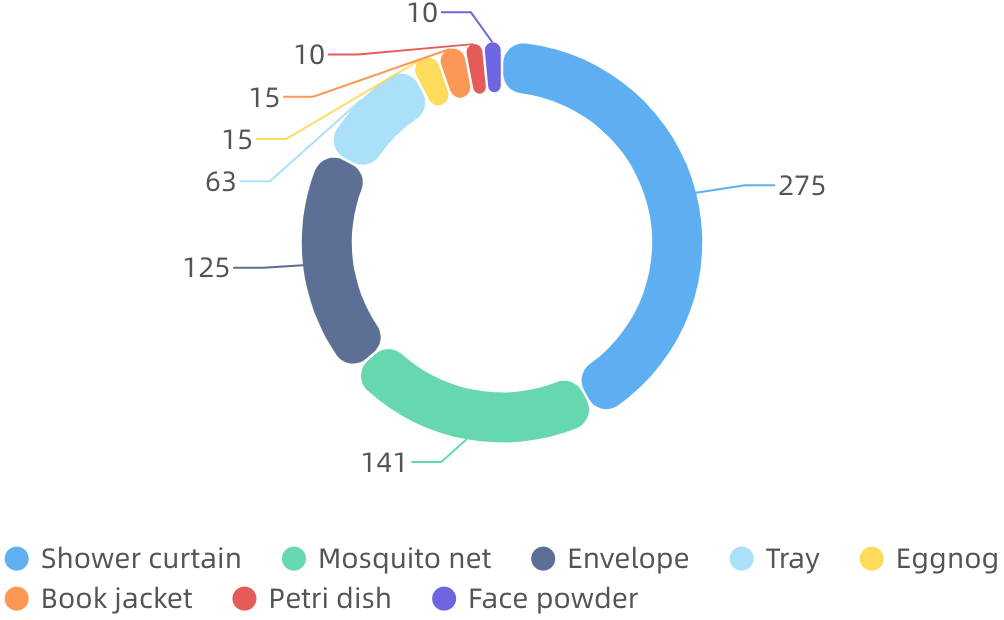}
\caption{Statistics on misclassification of adversarial samples. Through statistical analysis of all the misclassification results, we can observe the semantic information categories that were most commonly associated with the adversarial neon beams.}.
\label{figure6}
\end{figure}

\section{Evaluation}
\label{sec4}
\subsection{Experimental setting}
We utilize ResNet50 \cite{ref40} as the target model in all of our experimental tests, including digital and physical tests. Similar to the approach used in AdvLB \cite{ref35}, we randomly selecte 1000 images from ImageNet \cite{ref47} that are correctly classified by ResNet50, with each image belonging to a different category, as our dataset for the digital test. In the physical test, the devices employed in our experiments are illustrated in Figure \ref{figure4}. The neon lamp can project various colors, such as red, green, blue, yellow, with four levels of intensity, ranging from 0.2 to 0.6. For the camera device, we use an iPhone6s, and it has been confirmed that the AdvNB is not affected by different camera devices. We use a maximum of 4 neon lamps for physical tests due to the limitation of the illuminated area. For all tests, we utilize attack success rate (ASR) as the metric to evaluate the effectiveness of AdvNB, which is defined as follows:

\begin{equation}
\label{eq:Positional Encoding}
\begin{split}
    &{\rm ASR}(X) = 1-\frac{1}{O}\sum_{i=1}^{O}F({label}_{i})\\
    &F({label}_{i})=
        \begin{cases}
        1 & {label}_{i} \in {L}_{pre} \\
        0 & otherwise
        \end{cases}
\end{split}
\end{equation}
where $O$ is the number of clean samples that can be correctly classified in the dataset $X$, ${label}_{i}$ represents the ground truth label of the $i-th$ sample, ${L}_{pre}$  is the set of all labels predicted under attacking.

\begin{figure}
\centering
\includegraphics[width=1\columnwidth]{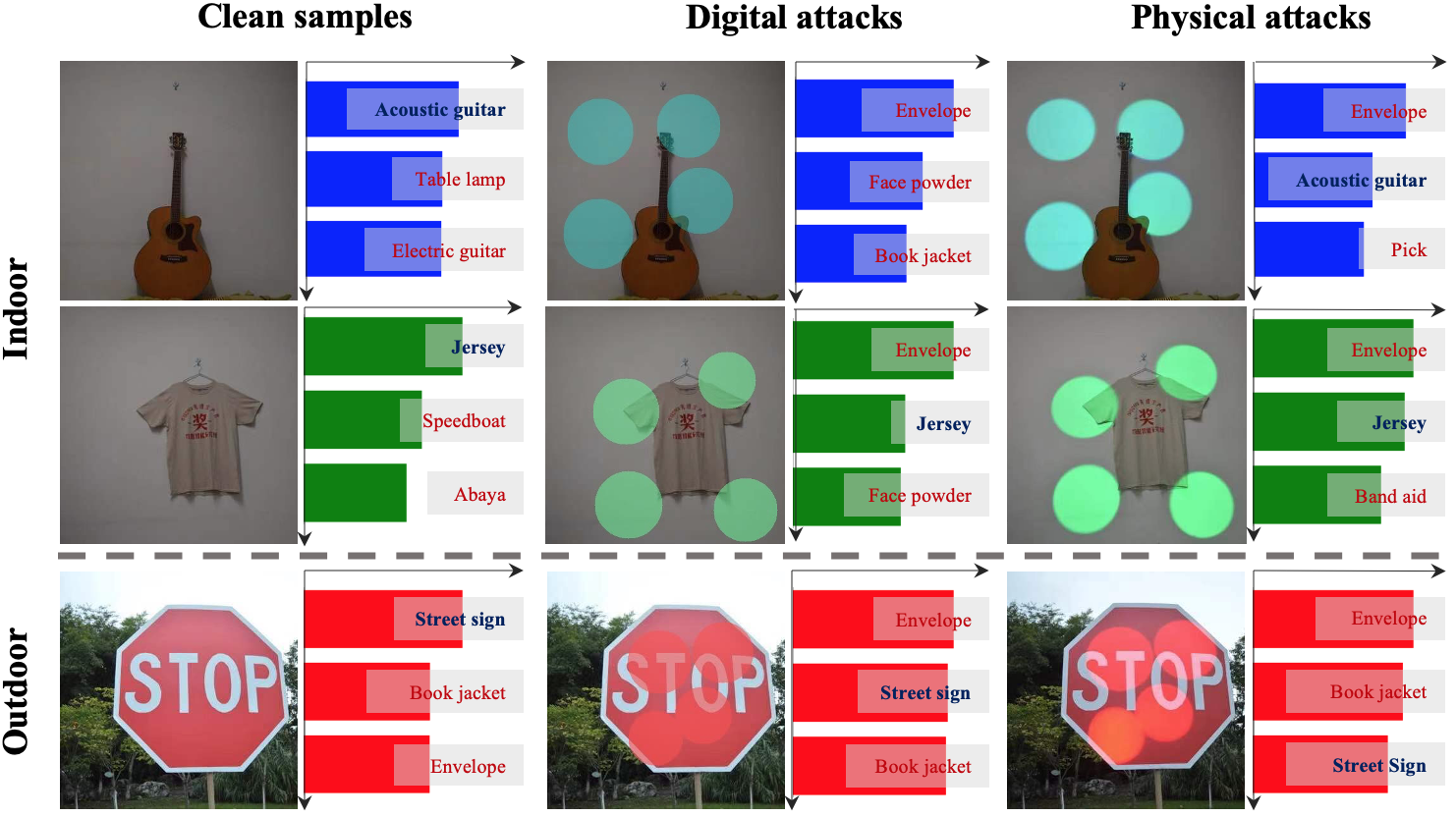}
\caption{Indoor/outdoor test. The first column corresponds to the clean samples, the second column corresponds to the generated digital samples, and the third column corresponds to the generated physical samples after irradiation with neon beams. The bar chart presents the TOP-3 prediction labels of the target model and the corresponding confidence comparison. As shown, the physical and digital samples exhibit a high degree of consistency. Additionally, the TOP-1 classification results of physical samples by the target model match those of digital samples.}.
\vspace{-0.4cm}
\label{figure7}
\end{figure}

\begin{figure*}
\centering
\includegraphics[width=1\linewidth]{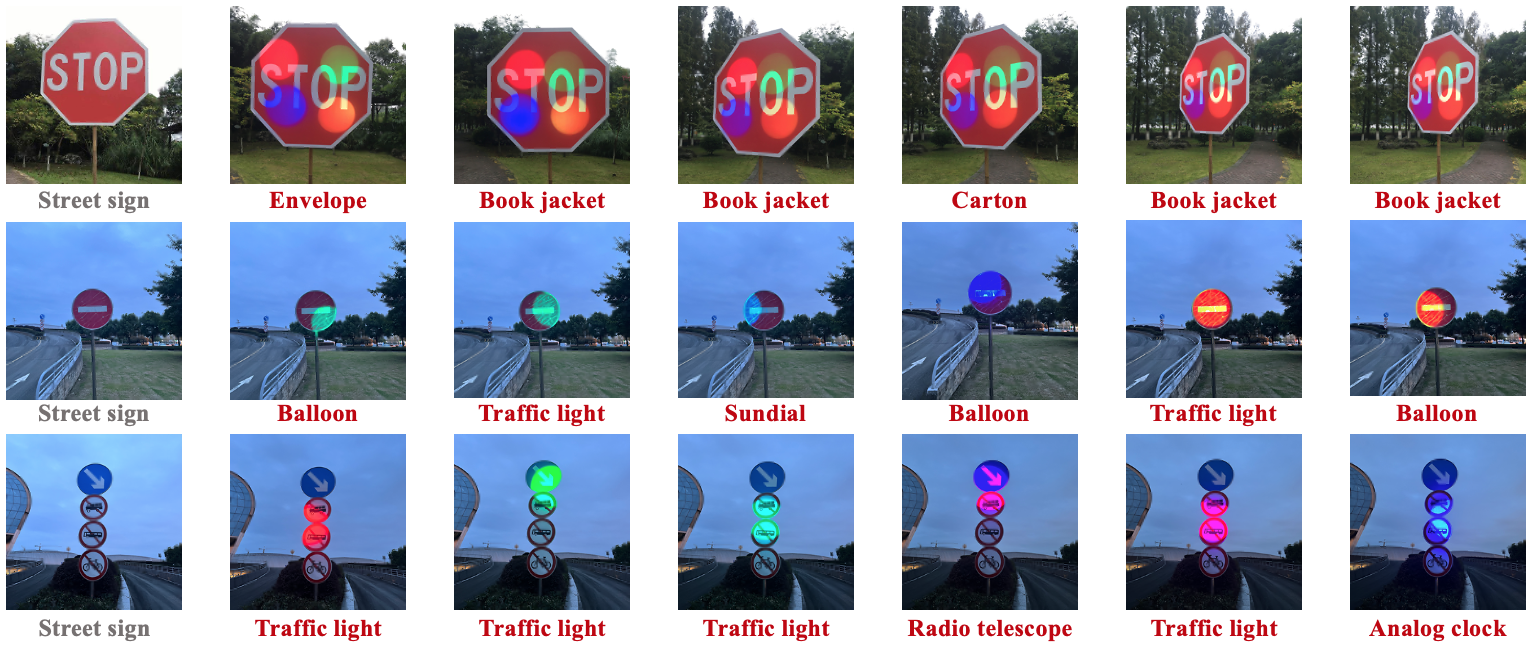}
\caption{Physical samples. The top row of the figure displays the physical samples obtained by attacking the stop sign using neon beams at angles of ${0}^{\circ}$, ${30}^{\circ}$, and ${45}^{\circ}$. In the second and third row, we use a single neon beam to attack the road signs, leading to incorrect classification by the classifier.}.
\vspace{-0.2cm}
\label{figure8}
\end{figure*}

\subsection{Evaluation of effectiveness}

\textbf{Digital test:} We conduct experiments in a digital setting using AdvNB on a dataset of 1000 images that were originally classified correctly by ResNet50. The results show an average ASR of 84.4\% with 189.7 queries, using 20 neon beams with $R=20$ pixels and $I=0.7$. More digital attack results can be found in the supplementary materials. Figure \ref{figure5} illustrates some interesting findings. The addition of neon beams of different colors to clean samples led the target model to misclassify the samples. For example, when neon beams are added to the "Seashore" image, the target model misclassify it as "Volcano", "Lakeside", "Balloon", and other categories. Although the adversarial neon beams in Figure \ref{figure5} are perceptible to human observers, the semantic information of the adversarial samples is consistent with that of the clean samples. In fact, some of the adversarial samples are so subtle that the perturbations are not even noticeable upon a quick glance. Furthermore, we perform statistical analysis on the misclassification results of adversarial samples. Figure \ref{figure6} shows that most of the digital samples are misclassified into a few categories such as "Shower curtain" and "Mosquito net". Therefore, we believe that the adversarial neon beam does not belong to any of the categories in ImageNet, but rather its semantic information is more similar to that of "Shower curtain" and "Mosquito net".

% In summary, AdvNB demonstrates a significant adversarial effect in the digital environment, as it causes advanced DNNs to misclassify images without altering their semantic content.

\textbf{Physical test.} We evaluate the efficacy of our proposed method in a physical setting, taking into account the potential impact of environmental noise on the effectiveness of physical attacks. To ensure the rigor of our experiments, we adopt a stringent experimental design that includes indoor and outdoor testing. The indoor test is designed to minimize the influence of outdoor noise, while the outdoor test aims to evaluate the performance of AdvNB in real-world scenarios.

% \begin{table}
%     \setlength{\abovecaptionskip}{0.3cm}%表格标题与表格间距太大
%     \setlength{\belowcaptionskip}{0.5cm}%表格标题与表格间距太大
%     \setlength{\belowdisplayskip}{5pt}
%     \centering
%     \caption{\label{Table 1}Attack success rates(ASR) from different angles.}
%     \begin{tabular}{cccccc}
%     \hline
%     Angle & ${0}^{\circ}$ & ${30}^{\circ}$ & ${45}^{\circ}$\\
%     \hline
    
%     AdvLB & 77.43\% & $\varnothing$ & $\varnothing$\\
%     \hline
%     AdvNB & 100.00\% & 100.00\% & 33.33\%\\
%     \hline
    
%     \end{tabular}
% \end{table}

\begin{table} 
	\centering
    \caption{Comparison of experimental results between AdvIB and AdvLB.}.
    \label{Table 1}
	\begin{tabular}{ccc|ccc}

    \hline
		\multirow{2}*{Method} & \multicolumn{2}{c}{Digital} & \multicolumn{3}{c}{Physical}\\
		\cline{2-6}
		~ & ASR(\%) & Query & ${0}^{\circ}$(\%) & ${30}^{\circ}$(\%) & ${45}^{\circ}$(\%) \\
		\hline
        AdvLB \cite{ref35}&\textbf{95.1}&834.0&77.4&$\varnothing$&$\varnothing$\\
        \hline
        AdvIB&84.4&\textbf{189.7}&\textbf{100}&\textbf{100}&\textbf{33.3}\\
        \hline

\end{tabular}
% \vspace{-0.3cm}
\end{table}

For the indoor test, we select 'Acoustic guitar', 'Jersey', and other objects as targets, and generate 36 adversarial samples, achieving a 100\% attack success rate (ASR of 100\% in AdvLB \cite{ref35}). The adversarial samples of the indoor test are presented in Figure \ref{figure7}, which shows that the computer-simulated neon beams maintain better consistency with the physical projected neon beams.
In the outdoor test, we select "Street sign" as the attack object and form 132 adversarial samples, achieving an attack success rate of 81.82\% (ASR of 77.43\% in AdvLB \cite{ref35}). Figure \ref{figure8} displays the adversarial samples generated in the outdoor environment. Attackers can use the adversarial neon beams, as shown in Figure \ref{figure8}, to execute physical attacks that lead advanced DNNs to misclassify "Street sign" as "Envelope", "Book jacket", etc. %To approach real scenarios, we conduct outdoor tests from different angles, and the experimental results are shown in Table \ref{Table 1}. The results demonstrate that AdvNB effectively launches physical attacks on target objects from various angles. Moreover, we analyze the misclassification of each angle. When the angle is 0 degrees, 91\% of the adversarial samples are classified as "Envelope"; when the angle is 30 degrees, 81\% of the adversarial samples are classified as "Book jacket"; and when the angle is 45 degrees, all adversarial samples are classified as "Book jacket".

We summarize the experimental results of AdvLB and our proposed method in Table \ref{Table 1}. As can be observed, in the digital environment, the ASR of our method is not superior to AdvLB, but our method exhibits significantly higher query efficiency than AdvIB. In the physical environment, our proposed method is capable of executing attacks from various angles, which is not feasible for AdvLB. Furthermore, the physical ASR of our method is higher than that of AdvLB. Overall, our approach outperforms the baseline methods.

% In conclusion, the results of our comprehensive experiments conducted in both digital and physical environments provide strong evidence for the effectiveness of AdvNB.

\subsection{Evaluation of stealthiness}
As previously stated, we choose neon beam as the physical perturbation in order to obtain a more natural physical sample, which makes our perturbation susceptible to being overlooked by human observers. As shown in Figure \ref{figure8}, our physical samples resemble natural neon beams falling on a street sign, and human observers have difficulty distinguishing between natural and artificial neon beams. On the other hand, the comparison of physical samples in Figure \ref{figure2} shows that the physical perturbation generated by AdvNB is more stealthy than the baseline. Given AdvNB's light-speed attack, AdvNB has greater temporal stealthiness than RP2 \cite{ref24} (RP2's physical perturbation will always adhere to the target object's surface, but AdvNB can control the light source, generating the physical perturbation only when the attack is carried out.). In contrast to AdvLB \cite{ref35}, the physical samples generated by AdvNB are more natural, allowing for better spatial stealthiness in our approach. When the cardboard is placed in front of the road sign for a shadow attack \cite{ref37}, it loses its spatial stealthiness, making human observers suspicious, whereas our method places the neon lamps far away from the target object, making AdvNB more stealthy than shadow attacks. In general, our approach results in a more stealthy attack than the baseline.

\subsection{Evaluation of robustness}
\textbf{Deploy AdvNB to attack advanced DNNs}.
We evaluate the robustness of the proposed AdvNB in a black-box setting with various classifiers, including the advanced DNNs (Inception v3 \cite{ref45}, VGG19 \cite{ref41}, ResNet101 \cite{ref40}, GoogleNet \cite{ref42}, AlexNet \cite{ref44}, MobileNet \cite{ref43}, DenseNet \cite{ref39}, Augmix+ResNet50 \cite{ref82}, ResNet50+RS \cite{ref83}, NF-ResNet50 \cite{ref84}). Note that the dataset is 1000 images selected from ImageNet that can be correctly classified by ResNet50, the physical parameters for the adversarial neon beams are configured as follows: $N$=20, $I$=0.7, $R$=20 pixels. Table \ref{Table 2} shows the ASR of our method with different classifiers. AlexNet is found to be the most vulnerable in the black-box attack test, with a 96.2\% ASR and an average of 128.3 queries. Furthermore, DNNs such as Augmix+ResNet50, ResNet50+RS, and NF-ResNet50 are more robust. In general, the data in Table \ref{Table 2} show that AdvNB have an adversarial effect of ASR on various models by more than 75\% in the black-box setting, confirming the robustness of our proposed AdvNB.

\begin{table}
\centering
\setlength{\abovecaptionskip}{0.5cm}
\caption{\label{Table 2} Evaluation across various classifiers.}
\begin{tabular}{cccc}
\hline

$f$ & Top-1 Accuracy(\%) & ASR(\%) & Query\\
\hline

Inception v3 & 87.6 & 82.5 & 196.4 \\
\hline

VGG19 & 91.5 & 89.6 & 157.2\\
\hline

ResNet101 & 96.1 & 81.3 & 209.7\\
\hline

GoogleNet & 85.3 & 85.5 & 184.6\\
\hline

AlexNet & 79.6 & 96.2 & 128.3\\
\hline

MobileNet & 89.7 & 86.2 &171.9\\
\hline

DenseNet & 90.8 & 87.1 &168.4\\
\hline

Augmix+ResNet50 & 93.7 & 79.8 & 221.6\\
\hline

ResNet50-RS & 94.6 & 77.2 & 229.5\\
\hline

NF-ResNet50 & 94.8 & 76.4 & 237.8\\
\hline

\end{tabular}
\end{table}

\begin{table}
\setlength{\abovecaptionskip}{0.3cm}%表格标题与表格间距太大
\setlength{\belowcaptionskip}{0.5cm}%表格标题与表格间距太大
\setlength{\belowdisplayskip}{5pt}
\centering
\caption{\label{Table 3}Transferability of AdvNB (ASR (\%)).}
\begin{tabular}{ccccc}

\hline
$f$ & Digital & Phy(${0}^{\circ}$) & Phy(${30}^{\circ}$) & Phy(${45}^{\circ}$)\\
\hline

Inception v3 & 75.9 & 77.8 & 39.5 & 4.5\\
\hline

VGG19 & 91.5 & 88.9 & 27.9 & 0\\
\hline

ResNet101 & 95.6 & 100 & 100 & 4.5\\
\hline

GoogleNet & 79.7 & 80.0 & 55.8 & 22.7\\
\hline

AlexNet & 95.1 & 100 & 100 & 79.5\\
\hline

MobileNet & 85.4 & 86.7 & 44.2 & 20.5\\
\hline

DenseNet & 76.9 & 82.2 & 51.2 & 20.5\\
\hline

Augmix+ResNet50 & 73.8 & 73.3 & 27.9 & 6.8\\
\hline

ResNet50-RS & 71.4 & 77.8 & 44.2 & 9.1\\
\hline

NF-ResNet50 & 70.2 & 75.6 & 41.9 & 6.8\\
\hline

\end{tabular}
\end{table}

\textbf{Transferability of AdvNB.}
Here, we present the attack transferability of AdvNB to advanced DNNs \cite{ref39,ref40,ref41,ref42,ref43,ref44,ref45,ref82,ref83,ref84} in both digital and physical settings. We use the adversarial samples generated by AdvNB that successfully attacked resnet50 as the dataset. The experimental results are illustrated in Table \ref{Table 2}. As shown, AdvNB exhibits excellent attack transferability in the digital environment, with an attack success rate above 90\% against VGG19, ResNet101, and AlexNet. In the physical environment, AdvNB demonstrates outstanding attack transferability at ${0}^{\circ}$, which effectively paralyzes most advanced DNNs. These results suggest that AdvNB can be utilized by attackers to exploit the transferability of AdvNB for efficient physical attacks against advanced DNNs, without any prior knowledge of the target model.

The experimental results in Table \ref{Table 2} and Table \ref{Table 3} demonstrate that AdvNB conducts robust physical attacks in a black-box setting. AdvNB enables attackers to carry out flexible operations, without any knowledge of the model, to perform effective physical attacks. Therefore, considering the remarkable adversarial effect of AdvNB on the vision-based system in real scenes, we advocate for the attention and further exploration of the proposed AdvNB.

\begin{figure}
\centering
\includegraphics[width=1\columnwidth]{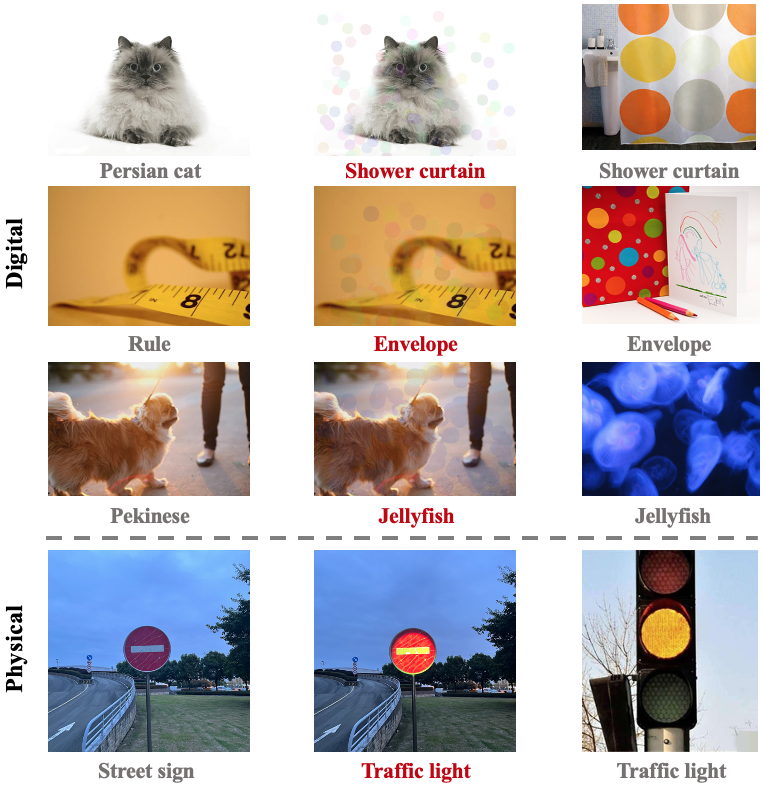}
\caption{Prediction errors caused by AdvNB. The results of tests conducted in both digital and physical environments demonstrate that the adversarial neon beam contains semantic information pertaining to certain categories of objects, which hinders the classifier's ability to accurately classify them.}.
\vspace{-0.3cm}
\label{figure10}
\end{figure}

\begin{figure*}
\centering
\includegraphics[width=1\linewidth]{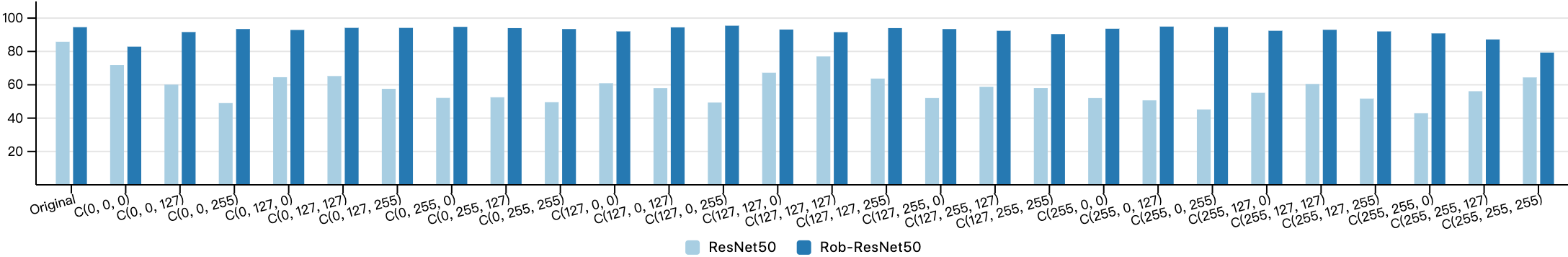}
\caption{Defense of AdvNB. It is evident that Rob-ResNet50 can achieve a TOP-1 classification accuracy of over 80\% for the 27 color adversarial samples, which is considered a basic level of adversarial defense. In fact, most defenses can achieve a TOP-1 classification accuracy of over 90\%.}.
\label{figure_Defense}
\end{figure*}

\section{Discussion}
\label{sec5}
\subsection{Analysis of prediction errors}
The neon beams act as adversarial perturbations, altering the features of the original images and providing new cues to DNNs. As previously stated, the majority of the digital samples are misclassified as "Shower curtain", "Envelope", and so on. Figure \ref{figure10} shows that the adversarial perturbations include several elements characteristic of a shower curtain. Additionally, our thorough analysis of the adversarial samples revealed some characteristic elements in "Envelope" that bear a strong resemblance to the perturbations. Moreover, single-color adversarial perturbations exhibit a significant similarity to "Jellyfish". In the physical experiment, we project a neon beam of color $C(255,255,0)$ onto a street sign, which the target model misclassified as a "Traffic light".

% \begin{figure*}
% \centering
% % \setlength{\belowcaptionskip}{-0.5cm}
% \includegraphics[width=1\linewidth]{figures/Ablation_C.png}
% \caption{Ablation of $C(r, g, b)$. To investigate the ablation experiment of $C(r, g, b)$ to the fullest extent possible, we perform a digital attack using a neon beam of 27 colors. It can be seen that most colors can achieve a higher ASR with a few queries. Only a few color attacks have low ASR and high queries, such as $C (127,127,127)$ and $C (0,0,0)$.}.
% \label{figure_AblationofC}
% \end{figure*}

% \begin{figure}
% \centering
% % \setlength{\belowcaptionskip}{-0.5cm}
% \includegraphics[width=1\columnwidth]{figures/fig9.png}
% \caption{Ablation of $N$, $R$ and $I$.}.
% \label{figure9}
% \end{figure}

\subsection{Defense of AdvNB}

In addition to showcasing the potential threats posed by AdvNB, we also explore potential defense strategies against this attack, particularly through the use of adversarial training. To this end, we construct a larger dataset in order to rigorously study the efficacy of our proposed defense strategy. Our dataset, which we call ImageNet-NeonBeam (ImageNet-NB), comprises 1.35 million adversarial samples generated by adding 27 different colors of simulated neon beams with $N$=20, $I$=0.7, and $R$=20 pixels to each of the 50 randomly selected clean samples from each of the 1000 categories in ImageNet \cite{ref47}.

We utilize the torchvision framework to train a robust ResNet50 model, which is optimized on 3 2080Ti GPUs using the ADAM optimizer with an initial learning rate of 0.01. As depicted in Figure \ref{figure_Defense}, the horizontal axis represents the various colors of neon beams added to the adversarial samples, and the vertical axis represents the corresponding TOP-1 classification accuracy. It is noteworthy that the robust model achieve a TOP-1 classification accuracy of over 90\% in most cases. We subject Rob-ResNet50 to AdvNB attack, where we use $N$=20, $I$=0.7, and $R$=20 pixels, resulting in 70.7\% ASR with an average query count of 353.8 (ResNet50: 84.4\% ASR, 189.7 average queries). These results indicate that although adversarial training can reduce AdvNB's ASR and increase AdvNB's attack time cost, it is not a foolproof defense against AdvNB. Please refer to the supplementary material for additional experimental results on Rob-ResNet50.

\subsection{Model attention}
\begin{figure}
\centering
\includegraphics[width=1\linewidth]{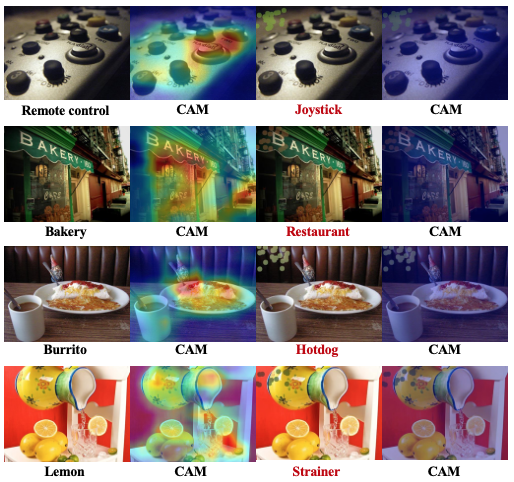}
\caption{CAM for images. }.
\vspace{-0.7cm}
\label{figure11}
\end{figure}

We utilize the Class Activation Mapping (CAM) technique \cite{ref48} to visualize the model's attention. In Figure \ref{figure11}, the first column displays the clean samples, while the third column displays the digital samples with the adversarial neon beam added to the upper left corner. A comparison of the second and fourth columns reveals that the model's attention is disrupted when the adversarial neon beam is added to the corner of the clean sample.

 % \subsection{Disadvantages of AdvNB}

 % In addition to introducing the effectiveness of our proposed method, we analyze some disadvantages of this method. Firstly, recent work such as AdvLB and OPAD has shown that they can only deploy outdoor attacks at nighttime and will be completely disabled during the daytime. Although AdvNB is one of the few light-based attacks that can be performed in daytime, the approach does not work in direct sunlight due to the limitations of current physical equipment. Secondly, in real scenario, neon beams scatter across the objects’ surfaces are smaller and denser. Our method cannot simulate neon beams in a way that human observers would consider it to be non-artificial. On the other hand, it can be seen from the defense experiment results that AdvNB is actually relatively easy to realize adversarial defense, which is an "imperfection" of the adversary. We will continue to improve AdvNB in our future work as well as improve our defense strategy against AdvNB.

\section{Conclusion}
\label{sec6}
In this work, we present a novel light-based physical attack called AdvNB that leverages the instantaneous nature of light to conduct effective physical attacks. Our evaluation criteria include effectiveness, stealthiness, and robustness. Extensive experimental designs and results demonstrate the effectiveness of AdvNB in both digital and physical environments. We showcase the stealthiness of our proposed method in terms of both temporal and spatial stealthiness by comparing the generated physical sample with a baseline. We employ AdvNB to launch attacks on advanced DNNs, demonstrate the attack transferability of AdvNB to verify its robustness. Our research highlights the security threat posed by light-based physical attacks to the physical world and sheds new light on future physical attacks using light as physical perturbations instead of stickers to enhance the flexibility of physical attacks. The proposed AdvNB, as an effective, stealthy, and robust light-based physical attack, provides a valuable complement to recent physical attacks.

In future work, we plan to apply the proposed AdvNB to suit various tasks, including object detection and domain segmentation. We also aim to explore other light-based physical attacks, such as adversarial reflected light. Additionally, developing effective defense strategies against light-based attacks will be a promising research direction.

\bibliographystyle{ieeetr}

\bibliography{IEEEfull}

% \bibliography{aaai22}

\end{document}